\documentclass{article}

\usepackage[a4paper, total={6in, 8in}]{geometry}

\usepackage{epsfig}
\usepackage{subfigure}
\usepackage{calc}
\usepackage{amssymb}
\usepackage{amstext}
\usepackage{amsmath}
\usepackage{amsthm}
\usepackage{pslatex}
\usepackage[small]{caption}

\usepackage[utf8]{inputenc}
\usepackage[T1]{fontenc}

\usepackage{authblk}

\newcommand{\xx}{\mathrm{\mathbf{x}}}
\newcommand{\yy}{\mathrm{\mathbf{y}}}
\newcommand{\XX}{\mathrm{\mathbf{X}}}
\newcommand{\CC}{\mathrm{\mathbf{C}}}

\usepackage{hyperref}  
\usepackage{graphicx}
\usepackage{algorithm}
\usepackage{algorithmic}
\usepackage{amssymb}

\begin{document}

\title{Model Guided Sampling Optimization for Low-dimensional Problems}

\author[1,2]{Lukáš Bajer}
\author[1]{Martin Holeňa}
\affil[1]{Institute of Computer Science, Academy of Sciences of the Czech Republic, Pod Vod\'{a}renskou v\v{e}\v{z}\'{i} 2, Prague}
\affil[2]{Faculty of Mathematics and Physics, Charles University in Prague, Malostransk\'{e} n\'{a}m. 25, Prague}

\onecolumn \maketitle \normalsize \vfill

\begin{center}
  ArXiv.org preprint \\
  original article published by SCITEPRESS -- Science and Technology Publications \\
  in \emph{ICAART 2015 Proceedings of the International Conference on Agents and Artificial Intelligence, Volume~2} \\
  ISBN 978-989-758-074-1
\end{center}

\abstract{
Optimization of very expensive black-box functions requires utilization of maximum information gathered by the process of optimization. Model Guided Sampling Optimization (MGSO) forms a more robust alternative to Jones' Gaussian-process-based EGO algorithm. Instead of EGO's maximizing expected improvement, the MGSO uses sampling the probability of improvement which is shown to be helpful against trapping in local minima. Further, the MGSO can reach close-to-optimum solutions faster than standard optimization algorithms on low dimensional or smooth problems. 
}

\section{\uppercase{Introduction}}
\label{sec:introduction}

\noindent
Optimization of expensive empirical functions forms an important topic in many engineering or natural-sciences areas. For such functions, it is often impossible to obtain any derivatives or information about smoothness; moreover, there is no mathematical expression nor an algorithm to evaluate. Instead, some simulation or experiment has to be performed, and the value obtained through a simulation or 
experiment is the value of the objective function being considered. Such functions are also called black-box functions. 
They are usually very expensive to evaluate; one evaluation may cost a lot of time and money to process.

Because of the absence of derivatives, standard continuous first- or second-order derivative optimization methods cannot be used. In addition, the functions of this kind are usually characterized by a high number of local optima where simple algorithms can be trapped easily. Therefore, different derivative-free optimization methods (often called meta-heuristics) have been proposed. Even though these methods are rather slow and sometimes also computationally intensive, the cost of the empirical function evaluations is always much higher and the cost of these evaluations dominates the computational cost of the optimization algorithm. For this reason, it is crucial to decrease the number of function evaluations as much as possible.

Evolutionary algorithms constitute a broad family of meta-heuristics that are frequently used for black-box optimization. Furthermore, some additional algorithms and techniques have been designed to minimize the number of objective function evaluations. All of the three following approaches use a model (of a different type in each case), which is built and updated within the optimization process.

\emph{Estimation of distribution algorithms} (EDAs) \cite{larranaga_estimation_2002} represent one such approach: EDAs iteratively estimate the distribution of selected solutions (usually solutions with fitness above some threshold) and sample this distribution forming a new set of solutions for the next iteration. 

\emph{Surrogate modelling} is the technique of learning and usage of a regression model of the objective function~\cite{jin_comprehensive_2005}. The model (called surrogate model in this context) is then used to evaluate some of the candidate solutions instead of the original costly objective function.

Our method, \emph{Model Guided Sampling Optimization} (MGSO), takes inspiration from both these approaches. It uses a regression model of the objective function, which also provides an error estimate. However, instead of replacing the objective function with this model,
it combines its prediction and the error estimate to get a probability of reaching a better solution in a given point. Similarly to EDAs, the MGSO then samples this pseudo-distribution\footnote{a function proportional to a probability distribution, it's value is given by the \emph{probability of improvement}}, 
in order to obtain the next set of solution candidates.

The MGSO is also similar to Jones' Efficient Global Optimization (EGO)~\cite{jones_efficient_1998}.
Like EGO, the MGSO uses a Gaussian process (GP, see~\cite{rasmussen_gaussian_2006} for reference), which provides a guide where to sample new candidate solutions in order to explore new areas and exploit promising parts of the objective-function landscape.
Where \emph{EGO selects} a single or very few solutions maximizing a chosen criterion -- Expected Improvement (EI) or Probability of Improvement (PoI) -- the \emph{MGSO samples} the latter criterion. 
At the same time, the GP serves for the MGSO as a surrogate model of the objective function for a small proportion of the solutions.

This paper further develops the MGSO algorithm introduced in~\cite{bajer_improving_2013}.
It brings several improvements (new optimization procedures and more general covariance function) and performance comparison to EGO. The following section introduces methods used in the MGSO, Section~\ref{sec:mgso} describes the MGSO algorithm, and Section~\ref{sec:results} comprises some experimental results on several functions from the BBOB testing set~\cite{hansen_real_2009}.

\section{\uppercase{Gaussian Processes}}

\noindent
Gaussian process is a probabilistic model based on Gaussian distributions. This type of model was chosen because it predicts the function value in a new point in the form of univariate Gaussian distribution: the mean and the standard deviation of the function value are provided. Through the predicted mean, the GP can serve as a surrogate model, and the standard deviation is an estimate of the prediction uncertainty in a new point.

The GP is specified by mean and covariance functions and a relatively small number of covariance function's hyper-parameters. The hyper-para\-meters are usually fitted by the maximum-likelihood method.

Let $\XX_N = \{\xx_i \ | \ \xx_i \in \mathbb{R}^{D}\}_{i=1}^{N}$ be a set of $N$ training $D$-dimensional data points with known dependent-variable values $\yy_N = \{y_i\}_{i=1}^{N}$ and $f(\xx)$ be an unknown function being modelled for which $f(\xx_i) = y_i$ for all $i \in \{1,\ldots,N\}$. The GP model imposes a probabilistic model on the data: the vector of known function values $\yy_N$ is considered to be a sample of a $N$-dimensional multivariate Gaussian distribution with the value of the probability density $p(\yy_N \, | \, \XX_N)$. If we take into consideration a new data point $(\xx_{N+1}, y_{N+1})$, the value of the probability density in the new point is
\begin{equation}
p(\yy_{N+1} \, | \, \XX_{N+1}) = \frac { \exp(-\frac{1}{2} \yy^\top_{N+1} \CC^{-1}_{N+1} \yy_{N+1}) } { \sqrt{(2\pi)^{N+1} \det(\CC_{N+1})} }
\end{equation}
where $\CC_{N+1}$ is the covariance matrix of the Gaussian distribution (for which mean is usually set to constant zero) and 
$\yy_{N+1} = (y_1,\ldots,y_N, y_{N+1})$ (see \cite{buche_accelerating_2005} for details). This covariance can be written as
\begin{equation}
\CC_{N+1} = \left( \begin{array}{cc} \CC_N & \mathbf{k} \\ \mathbf{k}^\top & \kappa \end{array} \right)
\end{equation}
where $\CC_N$ is the covariance of the Gaussian distribution given the $N$ training data points, $\mathbf{k}$ is the vector of covariances between the new point and training data, and $\kappa$ is the variance of the new point itself.

\paragraph{Predictions.} Predictions in Gaussian processes are made using Bayesian inference. Since the inverse $\CC^{-1}_{N+1}$ of the extended covariance matrix can be expressed using the inverse of the training covariance $\CC^{-1}_N$, and $\yy_N$ is known, the density of the distribution in a new point simplifies to a univariate Gaussian with the density
\begin{equation}
p(y_{N+1} \, | \, \XX_{N+1}, \yy_N) \ \varpropto \ \exp \left( -\frac{1}{2} \frac {(y_{N+1} - \hat{y}_{N+1})^2} {s^2_{y_{N+1}}} \right)
\label{univariate-density}
\end{equation}
with the mean and variance given by
\begin{eqnarray}
\hat{y}_{N+1} & = & \mathbf{k}^\top \CC^{-1}_N \yy_N, \\
s^2_{y_{N+1}} & = & \kappa - \mathbf{k}^\top \CC^{-1}_N \mathbf{k}.
\end{eqnarray}
Further details can be found in \cite{buche_accelerating_2005}.

\paragraph{Covariance functions.} The covariance $\CC_N$ plays a crucial role in these equations. It is defined by the covariance-function matrix $\mathbf{K}$ and signal noise $\sigma$ as
\begin{equation}
  \CC_N = \mathbf{K}_N + \sigma \mathbf{I}_N
\end{equation}
where $\mathbf{I}_N$ is the identity matrix of order $N$. Gaussian processes use para\-metrized covariance functions $K$ describing prior assumptions on the shape of the modelled function. The covariance between the function values at two data points $\xx_i$ and $\xx_j$ is given by $K(\xx_i, \xx_j)$, which forms the $(i,j)$-th element of the matrix $\mathbf{K}_N$. We used the most common squared-exponential function
\begin{equation}
K(\xx_i, \xx_j) = \theta \exp \left( -\frac{1}{2\ell^2} (\xx_i - \xx_j)^\top(\xx_i - \xx_j) \right), 
\label{eq:covseiso}
\end{equation}
which is suitable when the modelled function is rather smooth. The closer the points $\xx_i$ and $\xx_j$ are, the closer the covariance function value is to 1 and the stronger correlation between the function values $f(\xx_i)$ and $f(\xx_j)$ is. The signal variance $\theta$ scales this correlation,
and the parameter $\ell$ is the characteristic length-scale with which the distance of two considered data points is compared. Our choice of the covariance function was motivated by its simplicity and the possibility of finding the hyper-parameter values by the maximum-likelihood method. 

\section{\uppercase{Model Guided Sampling Optimization (MGSO)}}
\label{sec:mgso}

\noindent
The MGSO algorithm is based on a similar idea as EGO. It heavily relies on Gaussian process modelling, particularly on its regression capabilities and ability to assess model uncertainty in any point of the input space.

While most variants of EGO calculate new points from the \emph{expected improvement} (EI),
The MGSO utilizes the \emph{probability of improvement} which is closer to the basic concept of the MGSO: sampling a distribution of promising solutions\footnote{some EGO variants use PoI, too~\cite{jones_taxonomy_2001}}.

This probability is taken as a function proportional to a probability density and is sampled producing a whole population of candidate solutions -- individuals. This is the main difference to EGO which takes only very few individuals each iteration, usually the point maximizing EI.

\subsection{Sampling}
\label{sec:sampling}

\noindent
The core step of the MGSO algorithm is the sampling of the probability of improvement. This probability is, for a chosen threshold $T$ of the function value, directly given by the predicted mean $\hat{f}(\xx) = \hat{y}$ and the standard deviation $\hat{s}(\xx) = s_{y}$ of the GP model $\hat{f}$ in any point $\xx$ of the input space
\begin{equation}
  \mathrm{PoI}_T(\xx) = \mathrm{P}(\hat{f}(\xx) \leqq T) = \mathrm{\Phi}\left( \frac{T - \hat{f}(\xx)}{\hat{s}(\xx)} \right),
\end{equation}
which corresponds to the value of cumulative distribution function (CDF) of the Gaussian with density (\ref{univariate-density}) for the value of threshold $T$. As a threshold $T$, values near the so-far optimum (or the global optimum if known) are usually taken. 

\begin{figure}
  \centering
  \includegraphics[width=5cm]{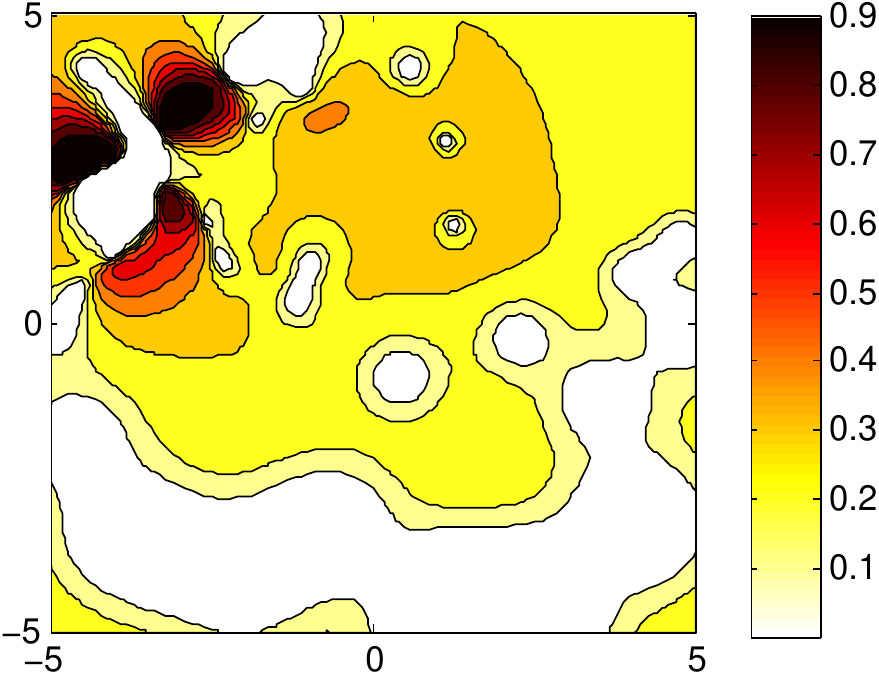}
  {\small
  \caption{Probability of improvement. Rastrigin function in 2D, the GP model is built using 40 data points.
  \label{fig:poi}
  }
  }
\end{figure}

Even though all the variables come from Gaussian distribution, $\mathrm{PoI}(\xx)$ is definitely not Gaussian-shaped since it depends on the threshold $T$ and the black-box function being modelled $f$. A typical example of the landscape of $\textrm{PoI}(\xx)$ in two dimensions for the Rastrigin function is depicted in Fig.\,\ref{fig:poi}.
The dependency on the black-box function also causes the lack of analytical marginals or derivatives.

\subsection{MGSO Procedure}

\noindent
The MGSO algorithm starts with an initial random sample from the input space forming an initial population, which is evaluated with the black-box objective function (step (2) in Alg.\,\ref{alg:mgso}). All the evaluated solutions are saved to a database from where they are used as a training set for the GP model.

\begin{algorithm}[t]
\begin{algorithmic}[1]
{\small
  \STATE \textbf{Input}: $f$ -- black-box objective function \\
      \quad $N$ -- standard population size \\
      \quad $r$ -- the number of solutions for dataset restriction
  \STATE $\mathsf{S}_0 = \{(\xx_j, y_j)\}_{j=1}^{N} \leftarrow$ generate $N$ initial samples 
      and evaluate them by $f$: \hspace{0.5em} $y_j = f(\xx_j)$
  \WHILE {no stopping condition is met, for $i=1,2,\ldots$}
    \STATE $\mathsf{M}_i \leftarrow$ build a GP model and fit its hyper-parameters
        according to the dataset $\mathsf{S}_{i-1}$
    \STATE $\{\xx_j\}_{j=1}^{N}  \leftarrow$ sample the $\textrm{PoI}_T^{\mathsf{M}_i}(\xx)$ 
        forming $N$ new points, optionally with different targets $T$ \\
    \STATE $\xx_\mathrm{min} \leftarrow \arg\min_{\xx} \hat{f}(\xx)$ -- find the minimum of the GP (by local cont. optimization) and replace the nearest solution from $\{\xx_j\}_{j=1}^{N}$ with it
    \STATE $\{y_j\}_{j=1}^N \leftarrow f(\{\xx_j\}_{j=1}^N)$  \hspace{\fill}  \COMMENT{evaluate the population} 
    \STATE $\mathsf{S}_i \leftarrow \mathsf{S}_{i-1} \cup \{(\xx_j, y_j)\}_{j=1}^N$  \hspace{\fill}  \COMMENT{augment the dataset} 
  \STATE $(\xx_\textrm{min}, {y}_\textrm{min}) \leftarrow \arg \min_{(\xx,y) \in \mathsf{S}_i} y$  \\
        \hspace{\fill}  \COMMENT{find the best solution in $\mathsf{S}_i$}
    \IF {any rescale condition is met} 
    \STATE restrict the dataset $\mathsf{S}_i$ to the bounding-box $[\mathbf{l}_i,\mathbf{u}_i]$ of the $r$ nearest solutions along the best solution $(\xx_\textrm{min},y_\textrm{min})$ and linearly transform $\mathsf{S}_i$ to $[-1, 1]^D$
    \ENDIF
  \ENDWHILE
  \STATE \textbf{Output}: the best found solution $(\xx_\textrm{min}, y_\textrm{min})$
}
\end{algorithmic}
\caption{MGSO (Model Guided Sampling Optimization)}
\label{alg:mgso}
\end{algorithm}

The main cycle of the MGSO starts with \emph{fitting} the GP model's ($\mathsf{M}_i$) hyper-parameters based on the data from the current dataset~$\mathsf{S}_i$ (step (4)). Further, the model's $\mathrm{PoI}_T^{\mathsf{M}_i}$ is \emph{sampled} (step (5)) and supplemented with the GP model's minimum (step (6)), forming up to $N$ new individuals $\{\xx_j\}_{j=1}^{N}$ where $N$ is a parameter defining the standard population size. The algorithm follows up with the evaluation of the new individuals with the objective function (step (7)), extending the dataset of already-measured samples (step (8)) and finding the best so-far optimum $(\xx_\textrm{min}, y_\textrm{min})$ (step (9)).

\paragraph{Covariance matrix constraint.} As every covariance matrix, Gaussian process' covariance matrix is required to be \emph{positive semi-definite} (PSD). This constraint is checked during sampling, and candidate solutions leading to close-to-indefinite matrix are rejected. Although it could cause smaller population-size in some iterations, it is an important step: otherwise, Gaussian process training and fitting becomes numerically very unstable. If such rejections arise, other two thresholds $T$
for calculating PoI are tested and population with the greatest cardinality is taken. 
These rejections occur when the GP model is sufficiently trained and sampled solutions become close to the model's predicted values.

\paragraph{Model minimum.} New population is supplemented with the minimum $\xx_\mathrm{min}$ of the model's predicted values found by continuous optimization\footnote{Matlab's \texttt{fminsearch} was used} (step (6), $\xx_\mathrm{min} = \arg\min_{\xx} \hat{f}(\xx)$); more precisely, the nearest sampled solution is replaced with this minimum (unless less than $N$ solutions were sampled).

\paragraph{Input space restriction.} In current implementation, MGSO requires bounds constraints $\xx \in [\mathbf{l}, \mathbf{u}], \, \mathbf{l} < \mathbf{u} \in \mathbb{R}^D$ to be defined on the input space, which is used by the algorithm to internal linear scaling of the input space to hypercube $[-1, 1]^D$. As the algorithm proceeds, the input space can be restricted along the so-far optimum to a smaller bounding box (steps (10)--(12)) which is again linearly scaled to $[-1, 1]^D$. The size of the new box is defined as a bounding box of $r$ nearest samples from the current so-far optimum $\xx_\textrm{min}$; enlarged by 10\% at the boundaries. For the parameter $r = 15 \cdot D$ was used as a rule of thumb.

This process not only speeds up model fitting and prediction (due to the smaller number of training data), but focuses the optimization along the best found solution and broaden small regions of non-zero PoI.

Several criteria are defined to launch such input space restriction, from which the most important is occurrence of numerous rejections in sampling due to close-to-indefinite covariance matrix. If the resulting new bounding box from the restriction is close to the previous box (the coordinates are not smaller than $[-0.8, 0.8]^D$), the input space restriction is not performed.

\subsection{Gaussian Processes Implementation}

\noindent
Our Matlab implementation of the MGSO makes use of Gaussian Process Regression and Classification Toolbox (GPML Matlab code) -- a toolbox accompanying Rasmussen's and Williams' monograph~\cite{rasmussen_gaussian_2006}. In the current version of the MGSO, Rasmussen's optimization and model fitting procedure \texttt{minimize} was replaced with \texttt{fmincon} from Matlab Optimization toolbox and with Hansen's Covariance Matrix Adaptation (CMA-ES)~\cite{hansen_completely_2001}. These functions are used for the minimization of GP's negative log-likelihood in model's hyper-parameters fitting. Here, \texttt{fmincon} is generally faster, but CMA-ES is more robust and does not need a valid initial point.

The next improvement lies in the employment of the diagonal-matrix characteristic length-scale parameter in the squared exponential covariance function, sometimes also called covariance function with automatic relevance determination (ARD)
\begin{multline}
  K^\mathrm{ARD}(\xx_i, \xx_j) = \\
  \theta \exp \left( -\frac{1}{2} (\xx_i - \xx_j)^\top \, \mathrm{diag}(\vec{\ell}) \, (\xx_i - \xx_j) \right).
\end{multline}
The length-scale parameter $\ell$ in (\ref{eq:covseiso}) changes to $\mathrm{diag}(\vec{\ell})$ -- a matrix with diagonal elements formed by the elements of the vector $\vec{\ell} \in \mathbb{R}^D$. The application of ARD was not straightforward, because hyper-parameters training tends to stuck in local optima. These cases were indicated by an extreme difference between the different scale-length parameter $\vec{\ell}$ components which resulted in poor regression capabilities. Therefore, constraints on maximum difference between components of $\vec{\ell}$ were introduced
\begin{displaymath}
  \vec{\ell}_i \ \leq \ 2.5 \ \| \textrm{median}(\vec{\ell}) - \vec{\ell}_i \|.
\end{displaymath}

\section{\uppercase{Experimental Results}}
\label{sec:results}

\begin{figure*}
  \centering
  \includegraphics[width=\linewidth]{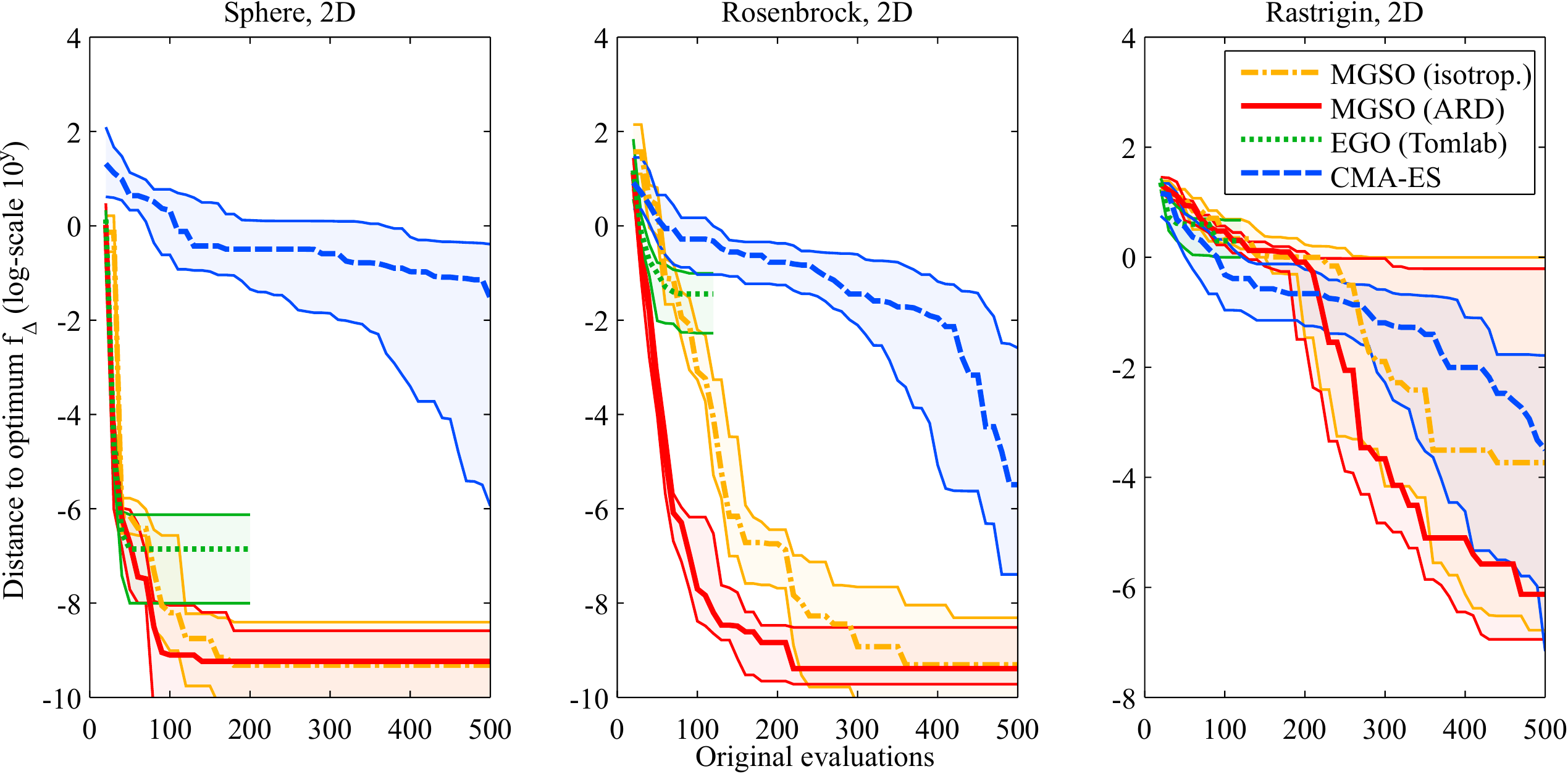}
  \vskip 1ex
  \includegraphics[width=\linewidth-3mm]{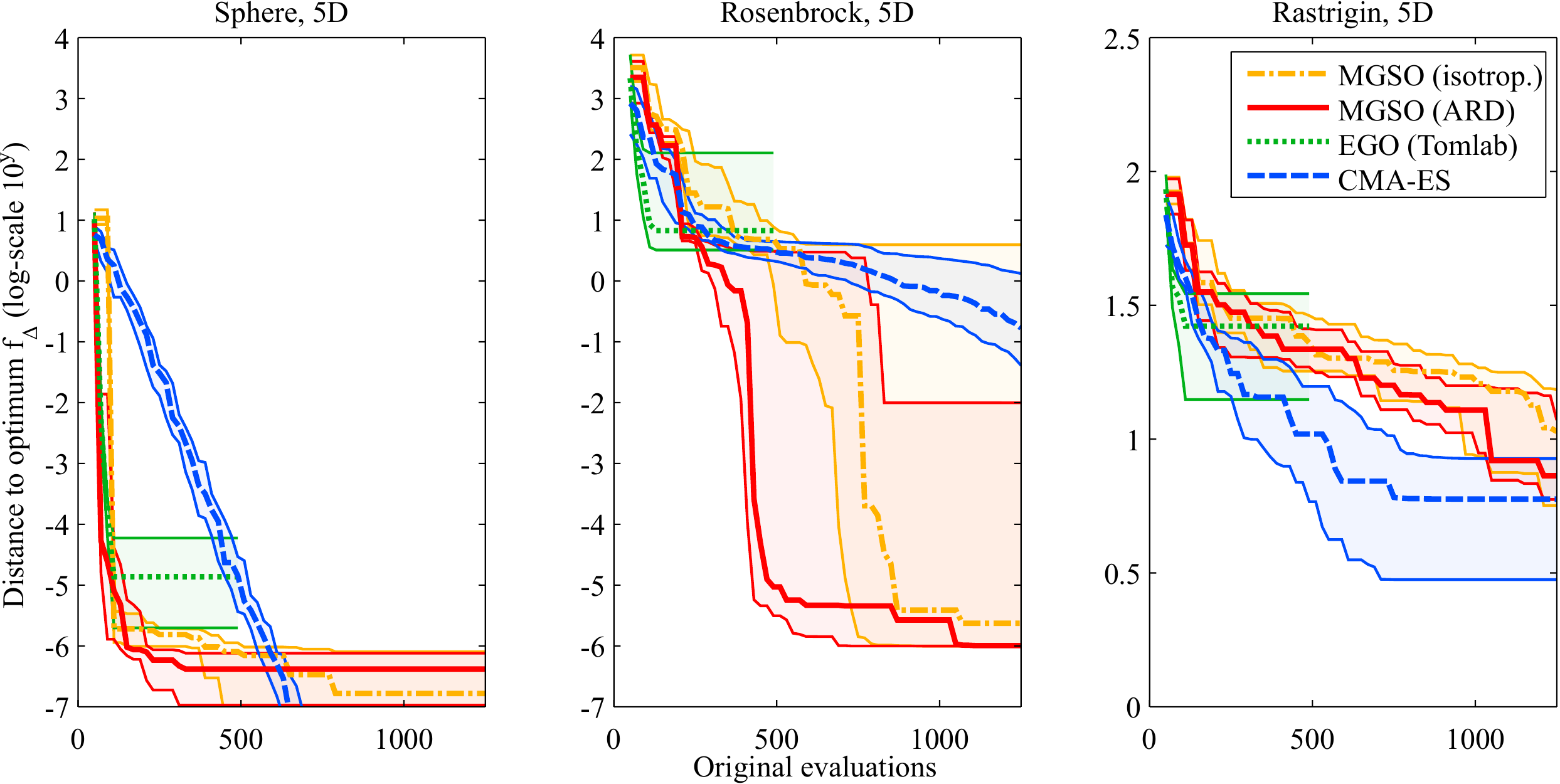}
  \vskip 1ex
  \includegraphics[width=\linewidth]{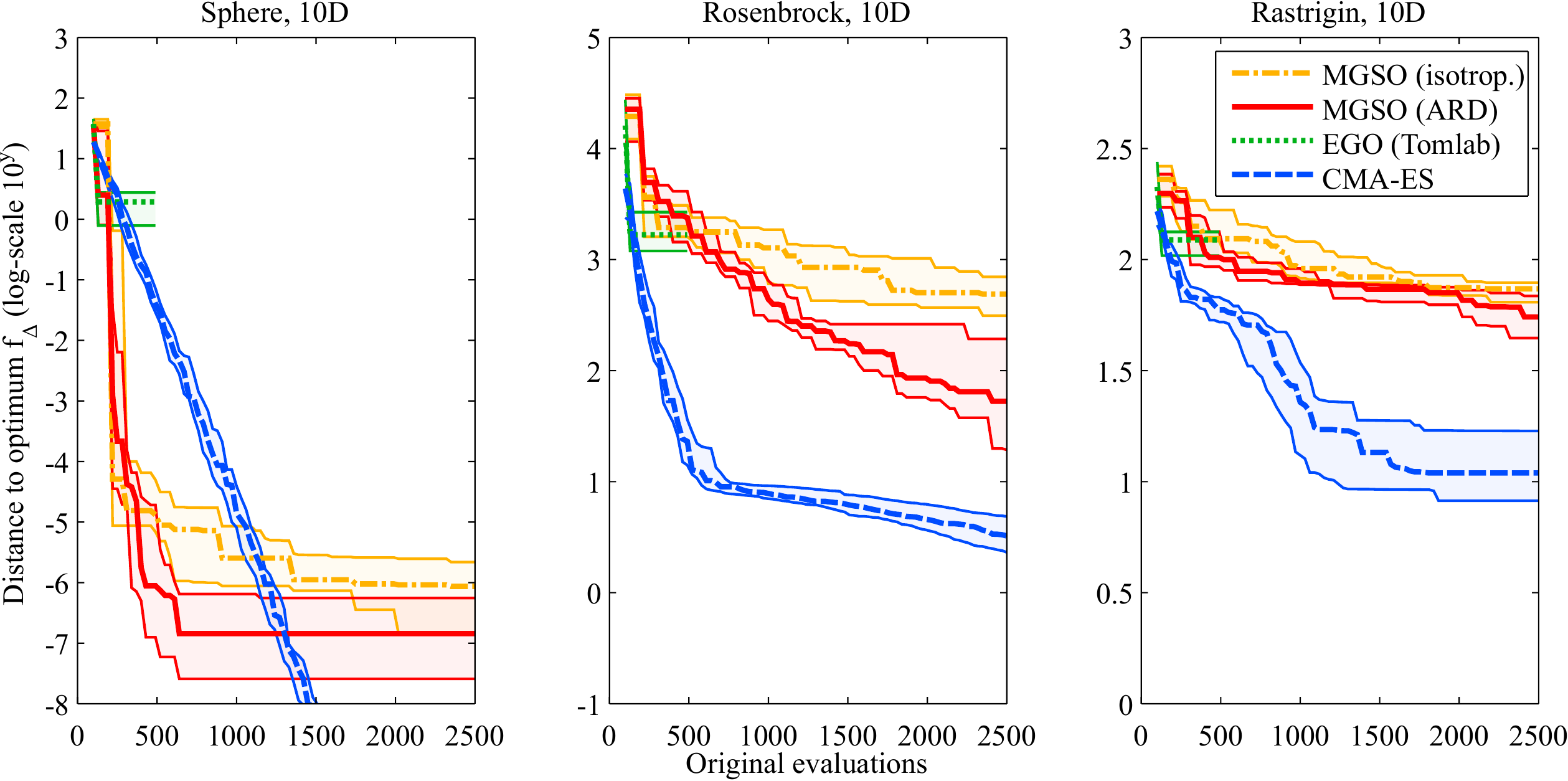}

  \caption{
    Medians, and the first and third quartiles of the distances from the best individual to optimum ($f_{\Delta} = f_\textrm{best} - f_\textrm{OPT}$) with respect to the number of objective function evaluations 
  for three benchmark functions. Quartiles are computed out of 15 trials with different initial settings (instances 1--5 and 31--40 in the BBOB framework).}
  \label{fig:test}
\end{figure*}

\noindent
This section comprises quantitative results from several tests of the MGSO as well as brief discussion of the usability of the algorithm. The current Matlab implementation of the MGSO algorithm\footnote{the source is available at \url{http://github.com/charypar/gpeda}} has been tested on three different benchmark functions from the BBOB testing set~\cite{hansen_real_2009}: sphere, Rosenbrock and Rastrigin function 
in three different dimensionalities: 2D, 5D and 10D. For these cases, comparison with CMA-ES -- current state of the art black-box optimization algorithm -- and Tomlab's implementation of EGO\footnote{http://tomopt.com/tomlab/products/cgo/solvers/ego.php} is provided.

The computational times are not quantified, but whereas CMA-ES performs in orders of tens of seconds, the running times of the MGSO and EGO reaches up to several hours. We consider this drawback acceptable since the primary use of the MGSO is expensive black box optimization where a single evaluation of the objective function can easily take many hours or even days and/or cost a considerable amount of money~\cite{holena_optimization_2008}.

In the case of \emph{two-dimensional problems}, the MGSO performs far better than CMA-ES on quadratic sphere and Rosenbrock function. The results on Rastrigin function are comparable, although with greater variance (see Fig.\,\ref{fig:test}: the descent of the medians is slightly slower within the first 200 function evaluations, but faster thereafter). The Tomlab's implementation of EGO performs almost equally well as the MGSO on sphere function, but on Rosenbrock and Rastrigin, the convergence of EGO is extremely slowed down after few iterations, which can be seen in 5D and 10D, too. The positive effect of ARD covariance function can be seen quite clearly, especially on Rosenbrock function. The difference between ARD and non-ARD results are hardly to see on sphere function, probably because its symmetry means no improvement in ARD covariance employment.

The performance of the MGSO on \emph{five-dimensional problems} is similar to 2D cases. The MGSO descends notably faster on non-rugged sphere and Rosenbrock functions, especially if we concentrate on depicted cases with a very low number of objective function evaluations (up to $250 \cdot D$ evaluations). The drawbacks of the MGSO is shown on 5D Rastrigin function where it is outperformed by CMA-ES, especially between ca. 200 and 1200 function evaluations.

Results of optimization in the case of \emph{ten-dimensional problems} show that the MGSO works better than CMA-ES only on the most smooth sphere function which is very easy to regress by Gaussian process model. On more complicated benchmarks, the MGSO is outperformed by CMA-ES. 

The graphs on Fig.\,\ref{fig:test} show that the MGSO is usually slightly slower than EGO in the very first phases of the optimization, but EGO quickly stops its progress and does not descent further. This is exactly what can be expected from the MGSO in comparison to EGO -- sampling the PoI instead of searching for the maximum can easily overcome situations where EGO gets stuck in a local optimum.

\noindent

\section{\uppercase{Conclusions and Future Work}}

\noindent
The MGSO, the optimization algorithm based on a Gaussian process model and the sampling of the probability of improvement, is intended to be used in the field of expensive black-box optimization. This paper summarizes its properties and evaluates its performance on several benchmark problems. Comparison with Gaussian-process based EGO algorithm shows that the MGSO is able to easily escape from local optima. Further, it has been shown that the MGSO can outperform state-of-the-art continuous black-box optimization algorithm CMA-ES in low dimensionalities or on very smooth functions. On the other hand, CMA-ES performs better on rugged or high-dimensional benchmarks.

\section*{\uppercase{Acknowledgements}}

\noindent This work was supported by the Czech Science Foundation (GA\v{C}R) grants \hbox{P202/10/1333} and \hbox{13-17187S}.

\bibliographystyle{apalike}
{\small
\bibliography{bajer2015icaart}
}

\end{document}